\documentclass{article}
\usepackage{spconf, amsmath, graphicx, url}
\usepackage{booktabs}
\usepackage[flushleft]{threeparttable}

\title{An Audio-Video Deep and Transfer Learning Framework for Multimodal Emotion Recognition in the wild}
\name{%
\begin{tabular}{@{}c@{}}
Denis Dresvyanskiy$^{1,2}$ \qquad 
Elena Ryumina$^{2,3}$ \qquad 
Heysem Kaya$^{4}$ \qquad 
Maxim Markitantov$^{3}$ \\
Alexey Karpov$^{3}$ \qquad 
Wolfgang Minker$^{1}$
\end{tabular}
}%
   
  \address{$^{1}$ Ulm University, Ulm, Germany\\
   $^{2}$ ITMO University, St. Petersburg, Russia\\
   $^{3}$ St. Petersburg Federal Research Center of the Russian Academy of Sciences, Russia\\
   $^{4}$ Department of Information and Computing Sciences, Utrecht University, Utrecht, The Netherlands}

\begin{document}
%
\maketitle
\begin{abstract}
In this paper, we present our contribution to ABAW facial  expression challenge. We report the proposed system and the official challenge results adhering to the challenge protocol. Using end-to-end deep learning and benefiting from transfer learning approaches, we reached a test set challenge performance measure of 42.10\%.
\end{abstract}
\section{Introduction}
\label{sec:intro}
Emotions play a vital role in daily human-human interactions and automated recognition of emotions from multi-modal signals has attracted increasing attention in the last decade with applications in domains ranging from  intelligent call centers to intelligent tutoring systems. Emotion recognition is studied in the broader affective computing field, where the studies of emotion is the focal point. The research in the domain is shifting to more ``in-the-wild", namely out of lab-controlled studies, thanks to new and challenging datasets collected and introduced over competitions such as Affective Facial Expressions in the Wild (AFEW)~\cite{dhallAFEW12,dhall2017individual} and Affective Behavior Analysis in the Wild (ABAW)~\cite{kollias2017recognition, zafeiriou2017aff, kollias2019deep, kollias2018multi, kollias2018aff, kollias2019expression}.

Motivated from the recent outstanding general performance of deep learning on audio and video domains as well as the efficacy of deep transfer learning to alleviate data scarcity in the target problem~\cite{kaya2017imavis,kaya2017multi,kollias2019deep}, in this study we employ both deep end-to-end learning and deep transfer learning for both audio and video modalities, fusing the scores of the uni-modal sub-systems for multi-modal affect recognition in out-of-lab conditions. We experiment on and use the official challenge protocol for the ABAW challenge - Facial Expressions Sub-challenge, originally run for Face and Gesture 2020, but was later extended until October. This sub-challenge features Ekman's six basic emotions plus neutral, thus featuring a seven-class classification task.

The contributions of this paper include 1) a novel multi-modal framework that leverages deep and transfer learning in audio and video modalities 2) extensive analysis of unimodal systems followed by a multimodal score fusion. 

\section{Proposed Approaches}
For the emotion recognition, we used both audio and video modalities. As a final solution, we introduce fusion system, which fuses probabilities from both modalities with different strategies. 

\subsection{Video-based deep networks}
As video-based models, we chose 2 same models in terms of structure, but different in terms of pretraining data were used.

\subsubsection{VGGFace2 based CNN}
\label{subsubsec:VGGFace2}
VGGFace2~\cite{cao2018vggface2} model was pretrained on the VGGFace2 dataset, which is intended to learn models recognize identities by their faces. It contains 8631 identities in training set with 362.6 images on average.
We took the pretrained model, cut the last layer provides class probabilities of the face, and stack two dense layers with 1024 and 7 neurons accordingly. The last layer allows the model to predict the probability of the emotional category.

To make the model more robust to interference, we used data augmentation, which is implemented via ImageDataGenerator included in keras~\cite{chollet2015keras} framework. In total, we used such augmentation techniques as rotation (up to 40 degrees), horizontal flip, and brightness (the value ranged from 0.2 to 1).  

Moreover, to decrease the effect of class disbalance on loss function, we applied logarithmic weighting. 

For training, the optimizer AdamW~\cite{AdamW} was chosen. The values of learning rate and weight decay were set to 0.00001. Also, the dropout with probability equaled 0.5 was applied to the last 2 layers. The model with the smallest loss value on the validation dataset across epochs was chosen. We used the predictions from this model called further non-temporal VGGFace2 as a first submission.

To capture temporal information across video frames, we implemented a 2D CNN + SVM model, which operates with windows of video. It was implemented in the following way: deep embeddings extracted from each frame by non-temporal VGGFace2 were formed in windows with specified size. Then, means and standart deviations (STDs) deep embeddings withing formed window were calculated. Thus, for every window we had 2048 features - 1024 means and 1024 STDs. We experimented with different sizes of windows and the value of best one equalled 4 seconds (it corresponds to 30 frames). As earlier, only one labels per whole window was chosen as a major category within corresponding window.

Next the Support Vector Machine (SVM) was trained on formed features. We conducted a lot of experiments with various kernels and regularization parameter and the best result was obtained with following parameters: kernel - \textit{poly}, \textit{gamma} = 0.1, \textit{C} = 3. It should be noted that we also applied logarithmic weighting, as earlier, but with parameter \textit{r} = 0.47. The predictions from this model were a third submission.

\subsubsection{VGG-FER CNN based Video Features}
Facial features are extracted over short non-overlapping video segments (e.g. 2s-4s) and summarized by statistical functionals. A deep neural network pre-trained with VGG-Face~\cite{Parkhi15} and fine-tuned with FER-2013 database~\cite{goodfellow2013challenges} is used. We use the aligned faces given by the ABAW challenge organizers to extract image-level deep features. The aforementioned transfer learning of VGG-Face based CNN by fine-tuning on FER-2013 is proposed in~\cite{kaya2017imavis}, and is successfully applied to a set of video based affect recognition challenges ranging from emotion recognition in the wild using AFEW corpora~\cite{dhallAFEW12} to apparent personality recognition~\cite{Gurpinar_2016_ICPR_Workshops, kaya2017multi} using ChaLearn LAP-First Impression corpus~\cite{Ponce_Lopez_LAPFIDB}.

The fine-tuned CNN can be reached over \textit{github repository\footnote[1]{\url{https://github.com/frkngrpnr/jcs}}}. The deep CNN has a 37-layer architecture (involving 16 convolution layers and 5 pooling layers). The response of the 33$^{rd}$ layer is used, which is the lowest-level 4\,096-dimensional descriptor. This output of the finetuned model is used  inline with the former works~\cite{kaya2017imavis,Gurpinar_2016_ICPR_Workshops,kaya2017multi} 
After extracting frame-level features from each aligned face using the deep neural network, non-overlapping short chunks of the original videos are summarized by computing mean and standard deviation statistics of each dimension over time. These features are subsequently modeled using Kernel Extreme Learning Machine (KELM), which is a fast and robust learning method~\cite{huang2012extreme}.

\subsection{Audio-based deep networks}

\subsubsection{Audio separation}
It is well-known that different extraneous sounds in audio recordings that are not speech (e.g. office, city, street noises, music) can significantly decrease the efficiency of the training process. To avoid it, we applied special separation library Spleeter~\cite{spleeter2020}, which contains a lot of pretrained deep neural models and is able to separate voice from other sounds. Since we wanted to separate just voice from all others sounds, we used a model, which allows us to divide audio on vocals (voice) and accompaniment (all other sounds including music). We downsampled original extracted audios to 16 kHz, because the Spleeter model is only able to work with no more then 16 kHz and then applied separation. Extracted vocals were used in further experiments.

\subsubsection{Labels preprocessing}
Because of the certain circumstances, different videos differ in their frame rate. As annotations were done in per-frame manner, they also have different sample rates. In order to equalize them, we down-sampled all labels to sample rate equals 5, since the smallest one frame rate is only 7.5 frames per second and the category of emotion switches very rarely. Thus, this value of the sample rate should be enough to catch all changes in emotions.

\subsubsection{PANN-based deep network}
Pretrained audio neural networks (PANNs)~\cite{kong2019panns} have demonstrated state-of-the-art performance in the audio pattern recognition domain. These models extract features from raw waveforms, a process that data, and return predictions in real-time. In this study, we used the CNN-14 model, which consists of one layer for extracting features, and six convolutional blocks, inspired by the VGG-like CNNs~\cite{simonyan2014very}. Each convolutional block consists of two convolutional layers with a kernel size of 3x3. Batch normalization is applied between each convolutional layer, and the ReLU nonlinearity is used to speed up and stabilize the training. Average pooling of a size of 2x2 is applied to each convolutional block for downsampling. After the last convolutional layer global pooling is applied to summarize the feature maps into a fixed-length vector.
CNN-14 model were preliminary trained on the large-scale AudioSet dataset~\cite{gemmeke2017audio}. We fine-tuned the PANN model for the Expression Challenge. We replaced the last fully-connected layer with a new one with seven neurons to get the probability of each category of emotion. All parameters are initialized from the CNN-14, except the final fully-connected layer which is randomly initialized. Raw waveforms with a window width of 3 seconds and a step of 1 second were fed to the input of the PANN model.

\subsubsection{1D CNN + LSTM based deep network}
Since there are no available pretrained 1D CNNs on raw audio to use it for transfer learning, we constructed and trained our own. Moreover, to catch temporal information from 1D CNN embeddings, we stacked two LSTM layers above it and seven neurons as a last layer of network to get the probability of each emotion category. In general, two schemes were realized:
\begin{itemize}
    \item Sequence-to-sequence modelling
    \item Sequence-to-one modeling (one emotion per whole window)
\end{itemize}

Thus, we implemented models be able to map input acoustic raw data into emotional category probabilities. The number of parameters of both models were equaled about 4.5 M. During the testing, the sequence-to-sequence model gave worse results, therefore in this work we present only sequence-to-one model.

As in video-based modeling, we conducted tests with different sizes of windows (from 2 to 12 seconds with step 2) and the best value was 4 seconds. To train final model, all audios were cut on parts of 4 seconds (with the intersection of 2/5 size of the window), which corresponds to 48000 values in waveform and 20 labels. The labels of window was selected as the most frequent category in it. 

The training process was conducted with following parameters: optimizer was Adam with \textit{learning rate} = 0.00025, loss function - \textit{categorical cross-entropy}, after every convolutional layer the dropout with rate equalled 0.3 was applied.

\subsection{Fusion}
\label{subsec:fusion}
We used two weighted score fusion methods. The first one assigns one weight for confidence scores of each model, where the weights sum up to 1. If we have $L$ models, then the optimized weights are a vector of length $L$. We use Dirichlet distribution to generate these weights randomly and try to optimize these weights with respect to the challenge measure on the validation set. We call this first approach ``Simple Weighted Fusion" (SWF).

In the second approach, we extend the first approach and have a fusion matrix of $L \times K$, where $K$ denotes the number of classes. That is we have an importance weight for each class of each model, separately. Similar to the SWF, the weights are randomly generated using Dirichlet distribution for each class, such that the sum of weights for each class over models sum up to unity. We call this ``Model and Class based Weighed Fusion" (MCWF). The latter approach has been successfully applied in former video based affect recognition in-the-wild challenges~\cite{kaya2015contrasting,kaya2017imavis}.

\section{Experimental Results}
The official performance measure used in challenge is defined by formula \ref{formula_metric}.
\begin{equation}\label{formula_metric}
    CPM = 0.67 * F_1 + 0.33 * Accuracy 
\end{equation}
where \textit{$F_1$} is a weighted average of the recall and the precision (also known as \textit{F}-measure) and the \textit{Accuracy} is the fraction of predictions that the model classified correctly.

We conducted extensive experiments with different unimodal video and audio systems separately and then fused them with different combinations. The results of experiments are presented in the Table~\ref{tab:valresults}.

As a baseline for our work, we took the results of non-temporal VGGFace2-based CNN, which was described in Section~\ref{subsubsec:VGGFace2}. We used the predictions from this model as a first submission. It reached a challenge performance measure of 50.23\% on the validation set and 40.60\% on the test set. 
For the subsequent submissions, to grasp the temporal dependencies we used VGGFface2 + SVM approach, which is also described in Section \ref{subsubsec:VGGFace2}. Since our second approach VGGFER-based CNN + KELM reached the result a little bit lower than the modified VGGFace2-based CNN, we decided to fuse three systems on the decision level. The fusion was done by SWF approach. The results of the fusion system are 56.56\% on the validation set and the predictions from this system were a second submission. On the test set the fused predictions showed 41.90\% of the challenge performance metric.

For the third submission, we chose predictions from modified VGGFace2-based CNN to check how it works standalone. The value of the competition metric of the model is 55.66\% on the validation set and 42.0\% on the test set.\\
For the next 3 submissions we used MCWF technique described in our paper with different combinations of models:

\begin{itemize}
    \item VGGFace2-based CNN + SVM and VGGFER-based CNN models fusion: the value of challenge performance metric is 55.70\% for validation set and 41.70\% for test set.
    \item VGGFace2-based CNN + SVM and 1D CNN + LSTM models fusion: the value of challenge performance metric is 55.87\% on validation set and 42.10\% on test set. It is the best performance value reached by our system in Expression challenge.
    \item VGGFace2-based CNN + SVM, 1D CNN + LSTM and PANN-based models fusion: the value of challenge performance metric is 54.78\% for validation set and 41.60\% for test set.
\end{itemize}
\begin{center}
\begin{table*}[ht] 

    \caption{The performance of models performed on ABAW challenge  (Expressions sub-challenge) validation set. \textbf{CPM}: Challenge performance measure (0.67 * F1-score + 0.33 * Accuracy). Baseline CPM=36.00\%}
    \label{tab:valresults}
	\centering
	\begin{threeparttable}
	\begin{tabular}{c c c c}
	\hline
    Features/Input & Model number & Validation CPM, \% & Test CPM, \% \\
	\hline
	Visual & 1 & 50.23 & 40.60 \\
	Visual & 2 & 55.66 & 42.00  \\
	Visual & 3 & 41.10 & -  \\
	Raw audio & 4 & 35.09 & -  \\
	Raw audio & 5 & 33.68 & - \\
	Visual & 2 \& 4 & 55.70 & 41.70 \\
	Visual & 2 \& 3 \& 4 & \textbf{56.56} & 41.90\\
	Visual \& Raw audio & 2 \& 4 & 55.87 & \textbf{42.10}\\
	Visual \& Raw audio & 2 \& 4 \& 5  & 54.78 & 41.60\\
	Visual \& Raw audio & 2 \& 3 \& 4 \& 5 & 55.54 & 41.80 \\
	\hline
    \end{tabular}
    
   \begin{tablenotes}
   \begin{center}
     \item[] \textit{\textbf{1}} - VGGFace2-based 2D CNN, \textit{\textbf{2}} - VGGFace2-based 2D CNN + SVM, \item[] \textit{\textbf{3}} - VGGFER-based 2D CNN \textit{\textbf{4}} -  1D CNN + LSTM, \textit{\textbf{5}} - PANN-based 2D CNN 
     \end{center}
   \end{tablenotes}
    \end{threeparttable}
\end{table*}
\end{center}

And, as a final submission, we again used SWF technique, but for all obtained models: VGGFace2-based CNN + SVM and VGGFER-based CNN features + ELM, PANN-based and 1D CNN + LSTM models.
The challenge performance metric on the validation set is 55.54\% and 41.80\% on the test set.

\section{Conclusions}
The paper has researched the efficiency of deep neural networks and the application of transfer learning and fusion techniques on emotion recognition problem. We have found that video-based MCWF fusion of different pretrained models with correctly fitted weights can increase the efficiency of the system compare to standalone models. However, audio-based models did not contribute to performance within the fusion process, which can be caused by information insufficiency of audios (the subjects often silence) or incorrectness of audio data such as music, a voice from another source, and noises. \\
In future work, it will be good to annotate audios manually for distinguishing different persons in audio and use semantic information extracted from a person's speech to add new modality in the fusion system.

\section{Acknowledgements}
The research was supported by the German Federal Ministry of Education and Research project "RobotKoop: Cooperative Interaction Strategies and Goal Negotiations with Learning Autonomous Robots". The experimental VGGFace2 research part was supported by the Russian Science Foundation (project No. 18-11-00145).


\bibliographystyle{IEEEbib}
\bibliography{mybib}

\end{document}